%% file: main.tex
\documentclass[10pt,twocolumn,letterpaper]{article}

\usepackage[pagenumbers]{cvpr} 

\usepackage{amsmath}
\usepackage{amsfonts}
\usepackage{multirow}

\input{preamble}
\definecolor{cvprblue}{rgb}{0.21,0.49,0.74}
\usepackage[pagebackref,breaklinks,colorlinks,allcolors=cvprblue]{hyperref}

\def\paperID{9179} 
\def\confName{CVPR}
\def\confYear{2026}

\title{ActVAR: Activating Mixtures of Weights and Tokens for Efficient Visual Autoregressive Generation}

\author{Kaixin Zhang$^1$, Ruiqing Yang$^2$, Yuan Zhang$^3$, Shan You$^4$, Tao Huang$^{5}$$\thanks{Corresponding author}$\\
$^1$School of Computer Science and Engineering, Central South University, \\
$^2$University of Electronic Science and Technology of China, \\
$^3$School of Computer Science, Peking University, \\
$^4$SenseTime Research, 
$^5$Shanghai Jiao Tong University\\
\tt\small kaixinzhang@csu.edu.cn, yrq@std.uestc.edu.cn, \\
\tt\small zhangyuan@alumni.pku.edu.cn, youshan@sensetime.com, t.huang@sjtu.edu.cn
}

\begin{document}
\maketitle
\input{sec/0_abstract}    
\input{sec/1_intro}
\input{sec/2_related_work}
\input{sec/3_proposed_approach}
\input{sec/4_experiment}
\input{sec/5_conclusion}
{
    \small
    \bibliographystyle{ieeenat_fullname}
    \bibliography{main}
}
\input{sec/6_suppl}

\end{document}

%% file: sec/0_abstract.tex
\begin{abstract}
Visual Autoregressive (VAR) models enable efficient image generation via next-scale prediction but face escalating computational costs as sequence length grows. Existing static pruning methods degrade performance by permanently removing weights or tokens, disrupting pretrained dependencies. To address this, we propose ActVAR, a dynamic activation framework that introduces dual sparsity across model weights and token sequences to enhance efficiency without sacrificing capacity. ActVAR decomposes feedforward networks (FFNs) into lightweight expert sub-networks and employs a learnable router to dynamically select token-specific expert subsets based on content. Simultaneously, a gated token selector identifies high-update-potential tokens for computation while reconstructing unselected tokens to preserve global context and sequence alignment. Training employs a two-stage knowledge distillation strategy, where the original VAR model supervises the learning of routing and gating policies to align with pretrained knowledge. Experiments on the ImageNet $256\times 256$ benchmark demonstrate that ActVAR achieves up to $21.2\%$ FLOPs reduction with minimal performance degradation.
\end{abstract}

%% file: sec/1_intro.tex
\section{Introduction}
\label{sec:intro}

Autoregressive models, which predict the next token based on preceding ones, have demonstrated remarkable success in image generation~\cite{van2017neural, ramesh2021zero, esser2021taming}. 
LLaMAGen~\cite{sun2024autoregressive} redesigns the tokenizer to support variable downsampling ratios, further improving image generation quality. 
Despite these advances, such methods remain limited by the standard next-token prediction paradigm, which typically requires a large number of generation steps and leads to slow inference.

Recently, VAR~\cite{tian2024visual} innovatively proposes the next-scale prediction paradigm, effectively enhancing image quality and generation speed. However, the VAR-based methods~\cite{tang2024hart, han2025infinity} approach introduces intermediate-scale features, which expand the sequence length and incur considerable computational cost.

\begin{figure}[t]
\centering
\includegraphics[width=0.95\columnwidth]{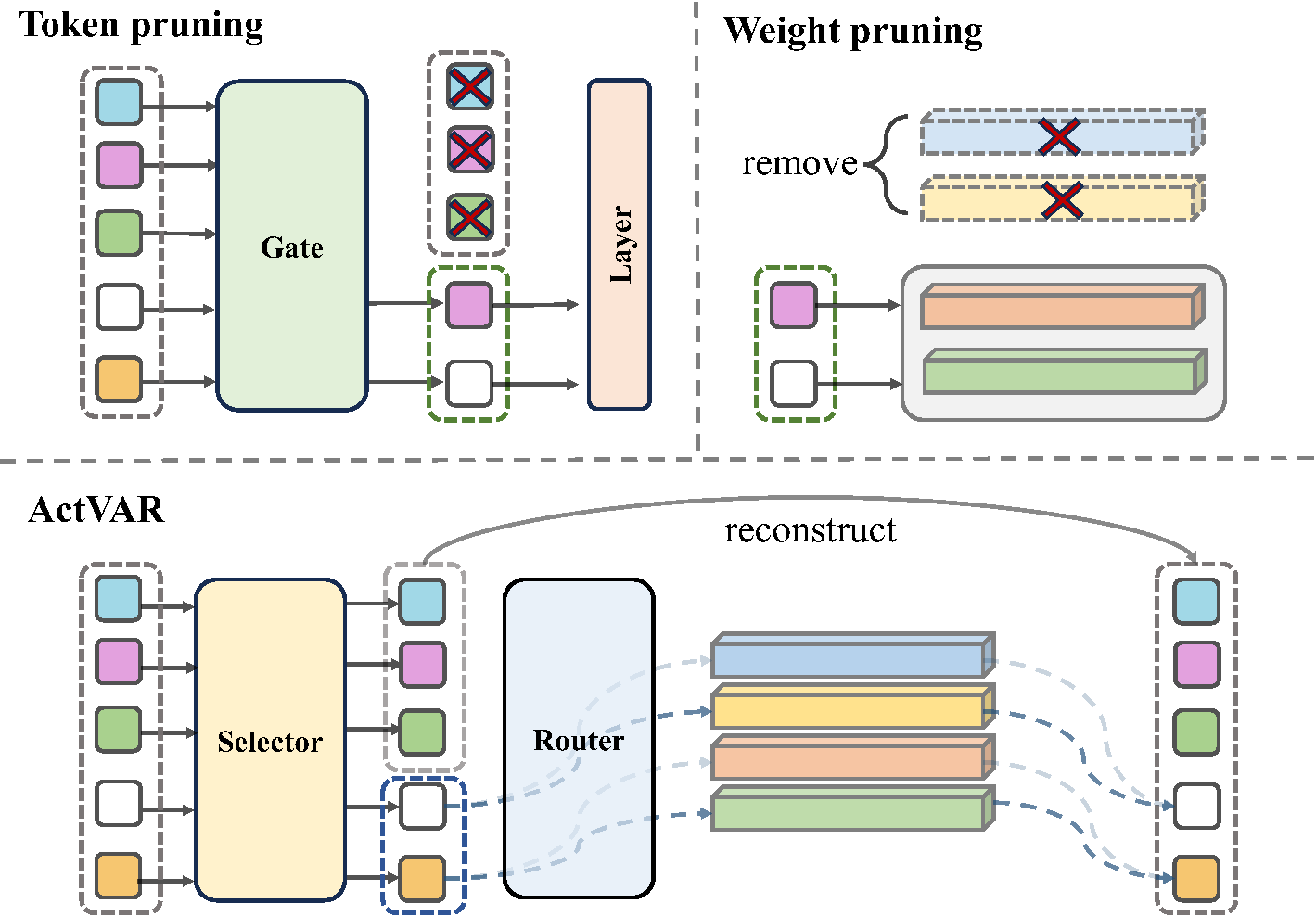}
\caption{\textbf{
Comparisons of conventional weight and token pruning methods and our proposed ActVAR.} Conventional methods permanently remove the weights and tokens, disrupting the dependencies. Our ActVAR achieves the same efficiency without sacrificing capacity.
}
\vspace{-4mm}
\label{fig:novel}
\end{figure}

\begin{figure*}[t]
\centering
\includegraphics[width=0.95\textwidth]{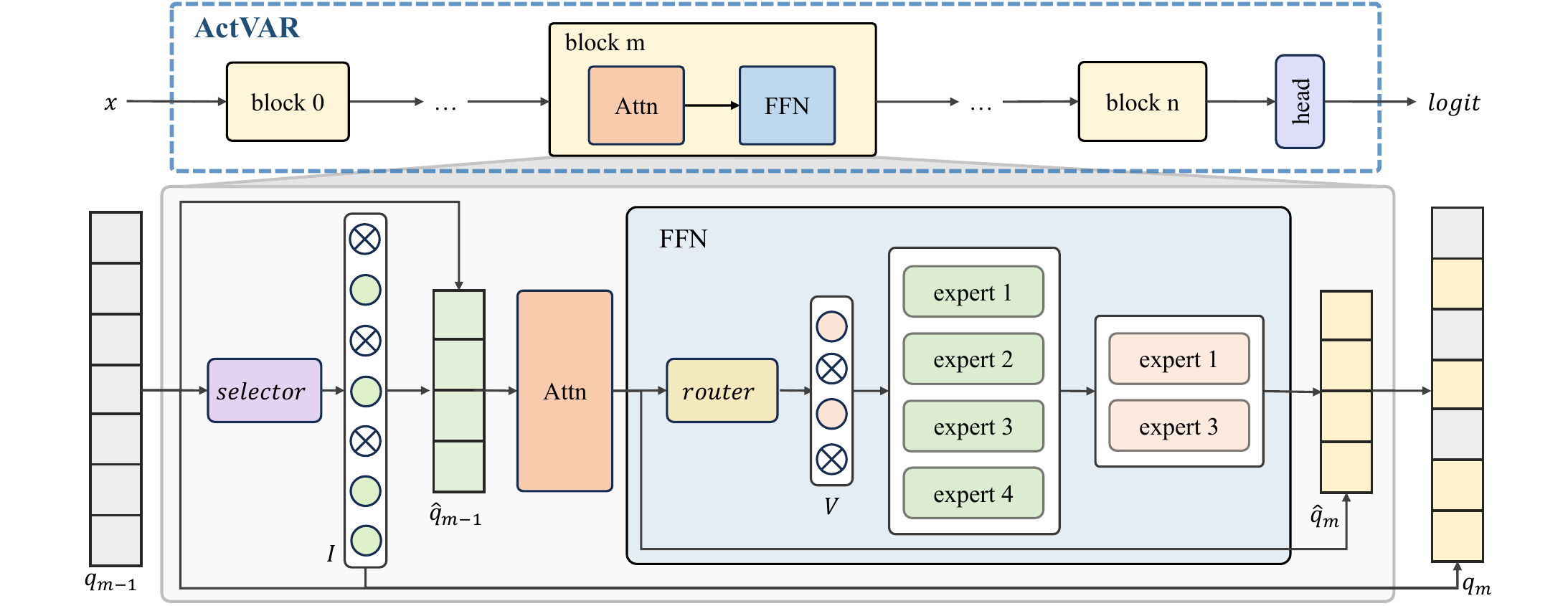}
\caption{\textbf{The pipeline of our ActVAR.} For the input sequence \( q_{m-1} \), the selector generates a binary indicator vector \( I \). Based on \( I \), a filtered input \( \hat{q}_{m-1} \) is constructed and passed through the attention and FFN layers. In the FFN, the router assigns each token to a subset of experts by predicting expert activation probabilities. Finally, the output \( \hat{q}_m \) is reconstructed into the complete sequence \( q_m \) using the indicator vector \( I \), maintaining alignment with the original scale.
}
\label{fig:framework}
\end{figure*}

A straightforward approach to reducing the computational cost of VAR models is to apply conventional weight and token pruning techniques for transformers (\emph{e.g.}, ToMe~\cite{bolya2022token} for ViT, and SparseVLM~\cite{zhang2024sparsevlm} for VLM). However, our empirical analysis reveals that such methods are only effective under low compression ratios\footnote{FastVAR yields merely the $8\%$ reduction in inference time on $256\times 256$ image generation, at the cost of increasing the FID from 2.29 to 2.64.}. We hypothesize that this limitation stems from the teacher-forcing training strategy employed in VAR, which makes the model particularly sensitive to structural changes. As illustrated in Figure~\ref{fig:novel}, the permanent removal of weights and tokens disrupts the learned dependencies and amplifies error accumulation during generation, ultimately degrading output quality. This motivates a key idea: \textit{can we dynamically activate only important weights and tokens at each layer, while keeping the rest unchanged, so that the model retains its full representational capacity and avoids information loss?}

In this paper, inspired by mixture-of-experts methods~\cite{zhou2022mixture, cao2025move}, we propose Activation VAR (ActVAR), a novel framework that introduces dual sparsity across both weights and tokens, as shown in Figure~\ref{fig:framework}. Specifically, we treat subsets of weights and tokens as experts, and dynamically activate only the most relevant subsets for each token and each layer. To achieve this, ActVAR decomposes the FFN into multiple sub-network experts and employs a learnable router to select token-specific expert combinations. To further reduce computation for long sequences, we introduce a gated token selector, which learns to identify and retain only the most up-to-date worthy tokens for each processing block. Crucially, unselected tokens are not discarded but incorporated into a lightweight reconstruction step that still preserves global context and sequence alignment for next-token prediction. 

For training ActVAR, we design a two-stage training strategy inspired by knowledge distillation~\cite{hinton2015distilling}, where a pre-trained VAR model acts as the teacher to supervise both the router and the selector. In the first stage, we train the router and gated token selector to adaptively assign experts and select informative tokens based on input semantics. In the second stage, we fine-tune the generative model to align with the learned routing and gating policies, while applying distillation losses to effectively transfer knowledge from the teacher model to ActVAR models.

In summary, the contributions of our work are as follows:

\begin{itemize}
    \item We propose \textbf{ActVAR}, a novel dynamic computation framework for autoregressive image generation that preserves the full model capacity while reducing inference overhead. Instead of permanently pruning weights or tokens, ActVAR selectively \textbf{activates} them at inference time based on input content, avoiding information loss and maintaining output quality.
    \item ActVAR introduces two lightweight routing modules: a \emph{weight router} that dynamically selects a subset of FFN sub-experts for each token, and a \emph{token router} that identifies important tokens to undergo further computation. This joint routing mechanism allows the model to adaptively allocate computational resources where needed.
    \item We conduct comprehensive experiments on the ImageNet $256\times 256$ benchmark, demonstrating that ActVAR achieves up to $21.2\%$ FLOPs saving with minimal performance drop.
\end{itemize}

%% file: sec/2_related_work.tex
\section{Related Work}
\label{sec:formatting}

\subsection{Autoregressive Models in Image Generation}
Prior works~\cite{reed2016generating, salimans2017pixelcnn++, lee2022autoregressive, zheng2022movq} introduce the next-pixel prediction paradigm, which adapts NLP-style token modeling to images by treating pixels as tokens. For example, some approaches~\cite{van2016conditional, chen2020generative, chen2018pixelsnail} typically flatten the two-dimensional image into a one-dimensional sequence using raster scan order, then employ CNNs or transformers to predict each pixel based on its predecessors. Subsequently, the tokenizer~\cite{van2017neural, esser2021taming} is introduced to compress continuous images into discrete tokens and use a transformer decoder to predict these tokens. Building upon this pipeline, many studies~\cite{sun2024autoregressive, wang2024omnitokenizer, zhu2024scaling, mattar2024wavelets} achieve performance on par with advanced diffusion models~\cite{rombach2022high, podell2023sdxl, peebles2023scalable}. However, the next-token prediction paradigm inherently requires a large number of decoding steps, which limits the overall efficiency of image generation. To address this, VAR~\cite{tian2024visual} recently proposed the next-scale prediction paradigm, which significantly reduces the number of generation steps and offers a promising direction for improving the efficiency of autoregressive models.

\begin{figure}[t]
\centering
\includegraphics[width=0.85\columnwidth]{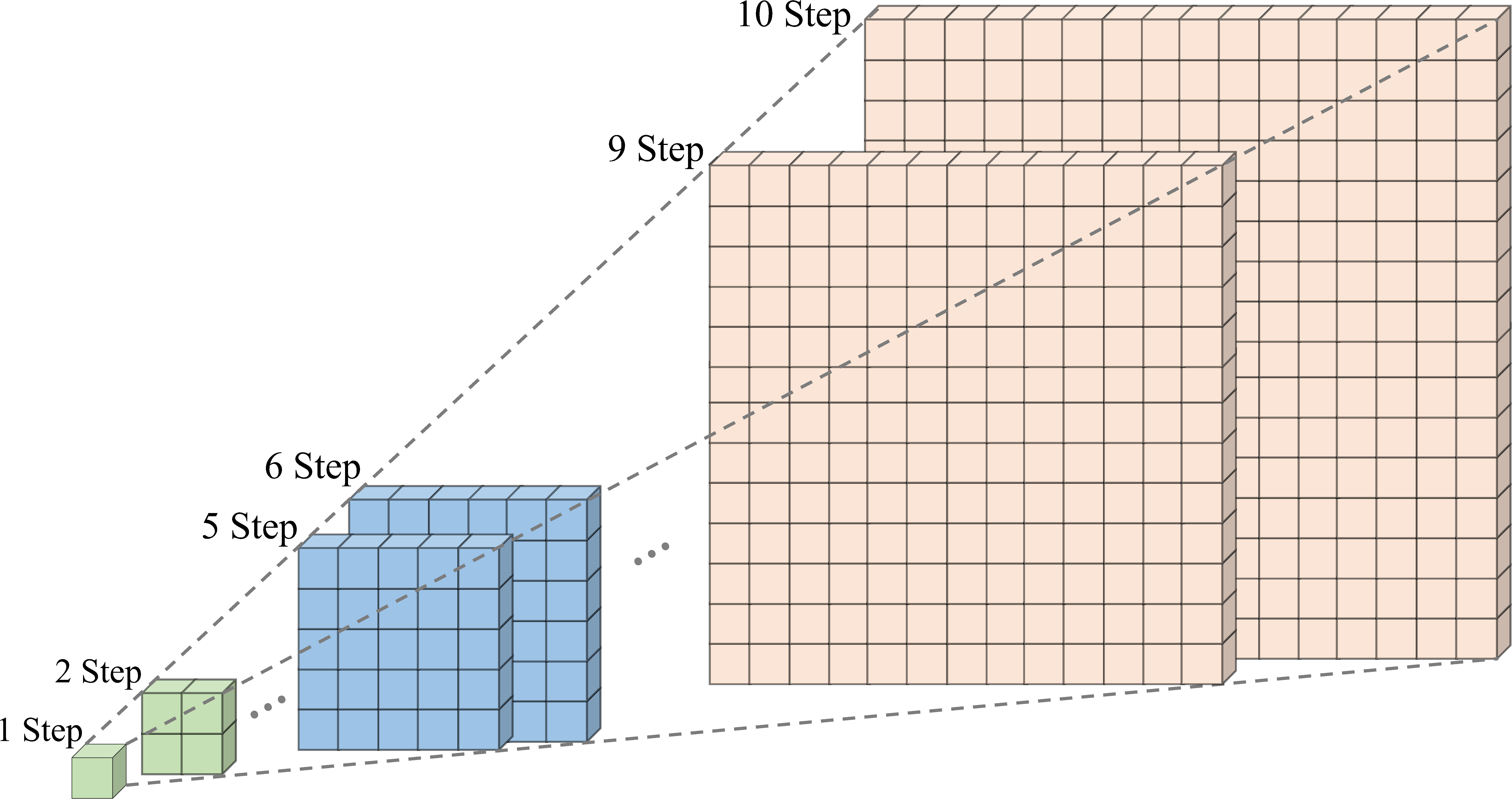}
\caption{\textbf{The number of tokens generated by VAR models at each step is increasing rapidly.}}
\label{fig:step}
\end{figure}

\subsection{Acceleration of Autoregressive Models}
Transformer~\cite{vaswani2017attention} has become a mainstream architecture in both natural language processing~\cite{brown2020language, devlin2019bert} and computer vision~\cite{dosovitskiy2020image, touvron2021training, liu2021swin}, largely due to their scalability and strong performance. However, this performance comes at the cost of substantial computational overhead, resulting in high resource demands for full-scale deployment. To address this issue, many studies~\cite{santacroce2023matters, ma2023llm, zhu2021vision, yu2023x} have proposed sophisticated techniques aimed at reducing model redundancy and accelerating inference. Inspired by the above studies, some efforts~\cite{li2024autoregressive, sun2024autoregressive, he2024zipar} focus on accelerating transformer-based autoregressive models. For example, the work~\cite{shen2025numerical} introduces structural weight pruning to compress the decoder-only transformer. Another study~\cite{anagnostidis2023dynamic} proposes a novel context pruning method to remove redundant tokens from the token sequence. Although VAR offers a promising alternative for efficient image generation, research on optimizing its computational performance remains scarce~\cite{guo2025fastvar, xie2024litevar}. This work aims to bridge that gap by investigating weights and token activation strategies tailored to VAR.

%% file: sec/3_proposed_approach.tex
\section{Proposed Approach: ActVAR}

\subsection{Computation Analysis of VAR}
Visual autoregressive (VAR) modeling introduces the innovative next-scale prediction paradigm, which replaces traditional token-by-token generation with the prediction of entire token maps at different scales. Given an image feature map $I \in \mathbb{R}^{H \times W \times C}$, VAR quantizes it into a set of multi-scale token maps $R = [r_1, r_2, \ldots, r_N]$. A transformer model is then used in an autoregressive manner to predict all token maps sequentially from small to large scales:
\begin{equation}
p(r_1, r_2, \ldots, r_N) = \prod_{i=1}^{N} p(r_i \mid r_1, r_2, \ldots, r_{i-1}),
\end{equation}
where the initial token $r_1$ is a class-condition embedding, and $N$ denotes the number of scales in total. 

During inference, VAR takes the interpolated upsampling of the last scale as input to predict all tokens at the next scale. This approach reduces the number of inference steps compared to vanilla autoregressive models. However, as shown in Figure~\ref{fig:step}, VAR still suffers from long input sequences at large scales, resulting in increased computational overhead. For instance, VAR requires a token sequence of length $680$ to generate a $256\times 256$ image, compared to only $256$ tokens in standard AR methods. As image resolution or generation granularity increases, this discrepancy in sequence length introduces substantial computational overhead, which severely constrains the scalability of vanilla VAR.

\begin{figure}[t]
\centering
\includegraphics[width=0.80\columnwidth]{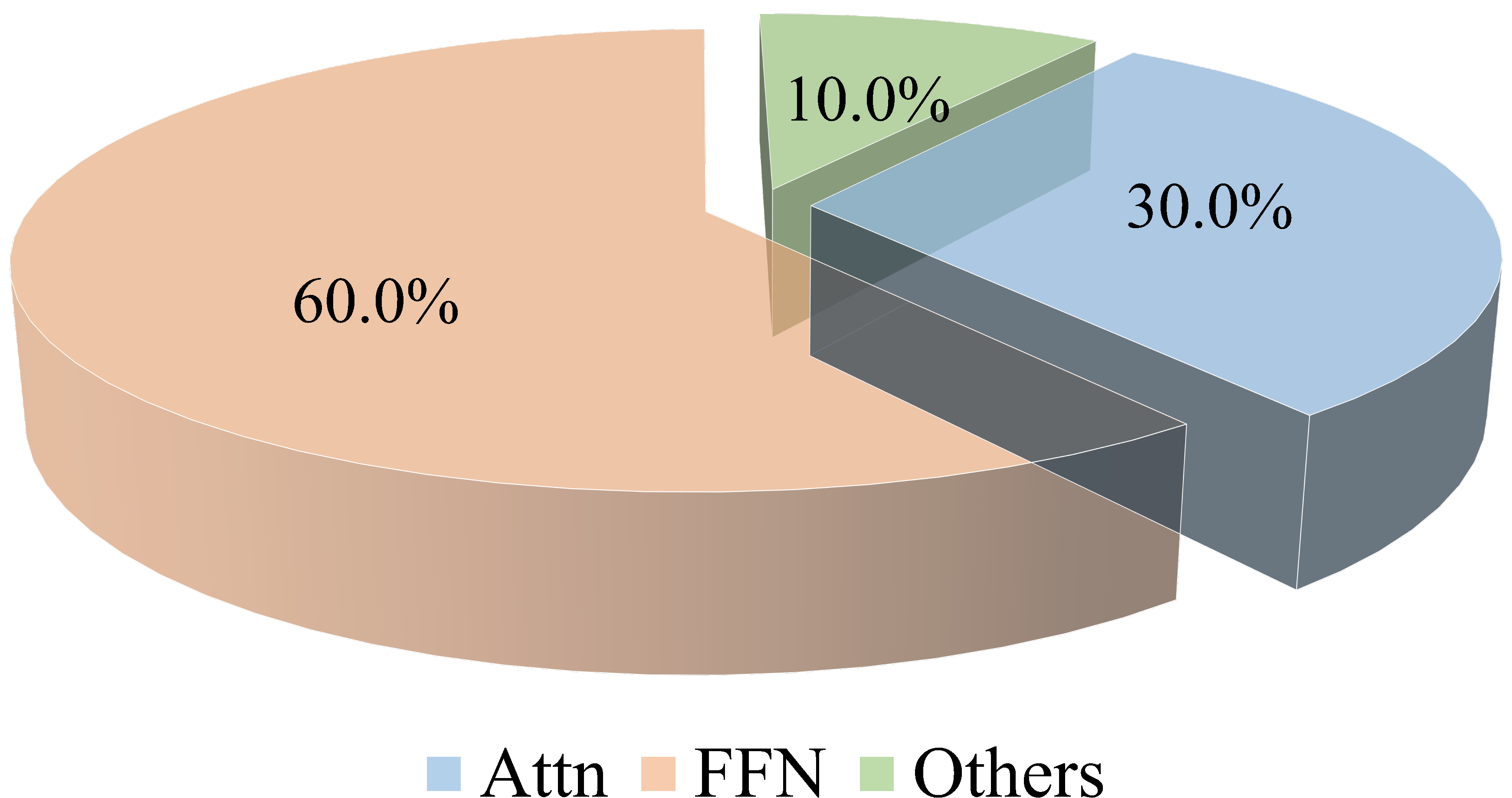}
\caption{\textbf{Distribution of computation per transformer block in VAR models.}}
\label{fig:flops}
\end{figure}

\begin{figure*}[t]
\centering
\includegraphics[width=0.95\textwidth]{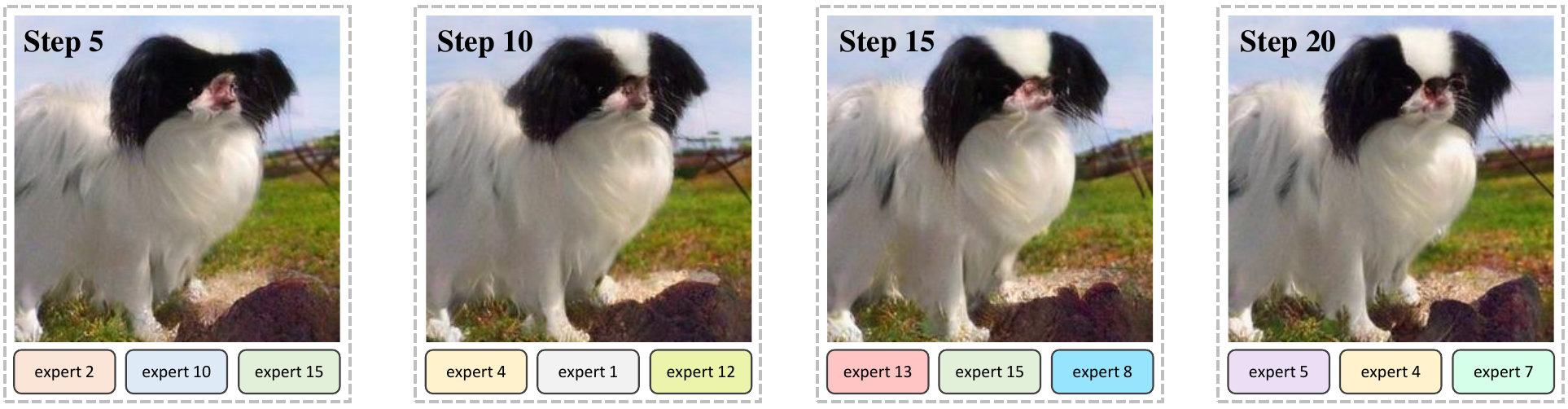} 
\caption{\textbf{Visualization of dynamic weight activations.} The FFN is partitioned into 16 expert networks. At a $16\times 16$ scale, the above four images show the top-3 expert networks for all token activations in different steps. Notably, the activation patterns vary significantly across steps, as tokens at different steps exhibit distinct preferences in their activation weights.}
\label{fig:weight_active}
\end{figure*}

To reduce the computational cost of VAR while keeping the generation capability as much as possible, we propose to dynamically activate only valuable computations and reduce the unnecessary computations as much as possible.

We begin by analyzing the computational costs of different modules in the VAR model. As shown in Figure~\ref{fig:flops}, most of the computations are introduced by the feedforward networks (FFNs) and attention layers. Besides, it is acknowledged that the FFNs are computationally expensive due to the high-dimensional dense projections, while the attention computation is proportional to the sequence length. As a result, to reduce both the computational costs of FFNs and attentions, we should consider reducing both the dimensions of FFNs and the sequence length of visual tokens.

As a result, we introduce ActVAR, a method to dynamically activate important weights and tokens. We will officially introduce our method as follows.

\subsection{FFN as Mixture of Experts}
The most direct way to improve the computational efficiency of FFNs is weight pruning. However, this method sacrifices model capacity and disrupts the original network structure. A more effective alternative is to retain the full set of weights and dynamically activate a subset based on the properties of the input during inference. As illustrated in Figure~\ref{fig:weight_active}, the preference activation weights dynamically change across steps while generating an image.

The linear layer, weighted by $W\in\mathbb{R}^{d_1\times d_2}$, can be regarded as the summation of $d_1$ matrix computations, \textit{i.e.},
\begin{equation}
    xW = \sum_i^{d_1} x_i W_{i}.
\end{equation}
This reminds us of mixture-of-experts (MoEs)~\cite{zhang2021moefication, zhao2024factorllm}, which adaptively select $K$ out of $N$ experts based on the input features. Therefore, we can use the idea of MoEs to selectively activate only a subset of FFN weights. Now we discuss the formal method for activating FFN weights as a mixture of experts.

\textbf{Weight decomposition.} The FFN module is implemented as a two-layer fully connected network. Given an input token $x$, the transformation process within the module can be described as follows:
\begin{equation}
\begin{split}
& h=xW_1 + b_1, \\
& F(x)=\delta(h)W_2 + b_2,
\end{split}
\end{equation}
where $W_1 \in \mathbb{R}^{d_p \times d_h}$ and $b_1 \in \mathbb{R}^{d_h}$ are the weights and biases of the first linear layer, and $W_2 \in \mathbb{R}^{d_h \times d_p}$ and $b_2 \in \mathbb{R}^{d_p}$ correspond to those of the second layer. The function $\delta(\cdot)$ denotes the activation function $GELU$.

In the construction of the expert networks, we divide the FFN into $N$ equally sized sub-networks. This design not only mitigates potential capacity bottlenecks but also eliminates latency issues that could arise from heterogeneous sub-network structures during parallel computation. 

Specifically, for a complete FFN, the decomposed parameters can be expressed as $W_1^i \in \mathbb{R}^{d_p \times d_e}$ and $W_2^i \in \mathbb{R}^{d_e \times d_p}$ for the weights, and $b_1^i \in \mathbb{R}^{d_e}$ for the biases, where the dimensionality of each sub-network satisfies $d_e = d_h / N$. Note that, according to the rules of matrix computation, the bias term $b_2$ remains unchanged.

\begin{figure*}[t]
\centering
\includegraphics[width=0.95\textwidth]{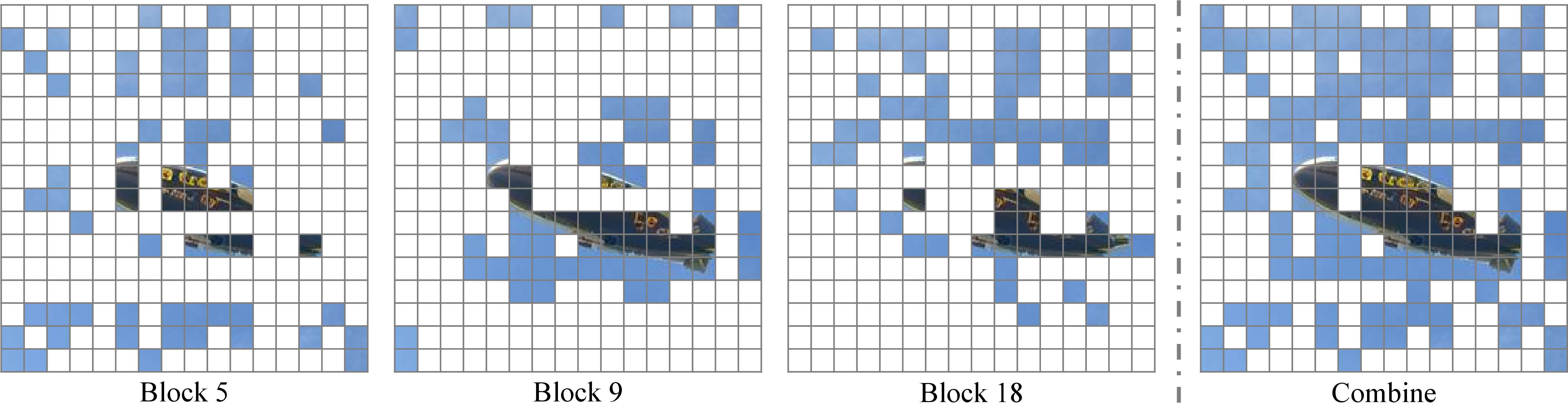} 
\caption{\textbf{Visualization of dynamic token activations.} The first three images on the left illustrate that, at a $16\times 16$ scale, each block dynamically activates tokens at distinct spatial positions. The image on the right aggregates the activation maps from the preceding three, highlighting the overall distribution of token activations.}
\label{fig:token_active}
\end{figure*}

\textbf{Dynamic activation of weights.} 
The routing mechanism is designed to adaptively assign a suitable expert network based on the characteristics of each input token. Specifically, for a given token $x$, the router $R(\cdot)$ computes the assignment probability $p_w \in \mathbb{R}^N$ for each expert network. The $\text{top-}k$ function is applied to $p_w$ to obtain the indicator vector $\mathcal{V} \in \{0, 1\}^N$ corresponding to the activated experts.

Based on the indicator vector, mapping rules are defined to construct the corresponding set of activated experts $\mathcal{E} = \{e^j\}_{j=1}^{K_w}$. 
Therefore, the final output of the dynamically activated experts can be formulated as the sum of the outputs from all selected experts:
\begin{equation}
        \sum_{j \in \mathcal{E}} e^j(x)
       = \sum_{j \in \mathcal{E}} \left( \delta(x W_1^j + b_1^j) W_2^j + b_2 \right),
\end{equation}
where $W_1^j$ and $W_2^j$ are the weights of the $j$-th expert. $b_1^j$ is the bias for the first layer, and $b_2$ is the shared bias term for the second layer.

\textbf{Optimize dynamic router.} Since the routing module lacks soft supervision, we need to construct an extra training signal. Inspired by knowledge distillation~\cite{hinton2015distilling, huang2022knowledge, zhang2024freekd}, we treat the original feedforward network $F_\theta$ as a teacher model, which incorporates comprehensive pre-trained knowledge and capabilities. Here, the expert networks $\mathcal{E}_{\widetilde{\theta}} = \{e^i\}_{i=1}^{N}$ are regarded as a group of student models. For each expert, we compute the mean squared error (MSE) between its output and that of the teacher, yielding a distance vector $\{d_i^{\text{w}}\}_{i=1}^N$. 
The distance is intended to quantify the divergence between the expert output and the teacher output.
Therefore, a top-$k$ selection based on the distances is performed to identify the most relevant experts and yields a pseudo-label matrix:
\begin{equation}
\mathcal{A}_w = \sigma(\text{top}-k(\{-d_i^{\text{w}}\}_{i=1}^N, K_w)),
\end{equation}
where $\sigma(\cdot)$ denotes the softmax function. Under the supervision of the pseudo-label matrix $\mathcal{A}_w$, we employ a distillation loss to encourage the routing module $R(\cdot)$ to align its predicted probabilities $p_w$ with the softened $\mathcal{A}_w$:
\begin{equation}
L_{dis}^w= \text{KL}(p_w \parallel \sigma(\mathcal{A}_w)))=\sum_{i=1}^{N}p_w\log\frac{p_w}{\sigma(\mathcal{A}_w)}, 
\end{equation}
where $\text{KL}(\cdot)$ is the Kullback–Leibler divergence. 

Furthermore, MoE-based architectures often suffer from load imbalance and degraded performance due to over-reliance on a small subset of experts~\cite{fedus2022switch}. To mitigate this issue, we follow~\cite{lepikhin2020gshard} and incorporate a load balancing loss that penalizes the uneven distribution of input tokens across experts:
\begin{equation}
L_{bl} = \frac{K_w}{N}\sum_{i=1}^N\sum_{j=1}^{K_w} \mathbb{I}_j(x_i)R_j(x_i),
\label{eq:balance}
\end{equation}
where $x_i$ denotes the $i$-th token in the input sequence. The indicator function $\mathbb{I}_j(x_i)$ equals $1$ if the $j$-th expert is selected to process $x_i$ via the top-$K$ routing mechanism, and $0$ otherwise. In this way, the routing mechanism can learn to mimic the implicit preference of the teacher model for expert specialization.

\begin{figure*}[t]
\centering
\includegraphics[width=1.0\textwidth]{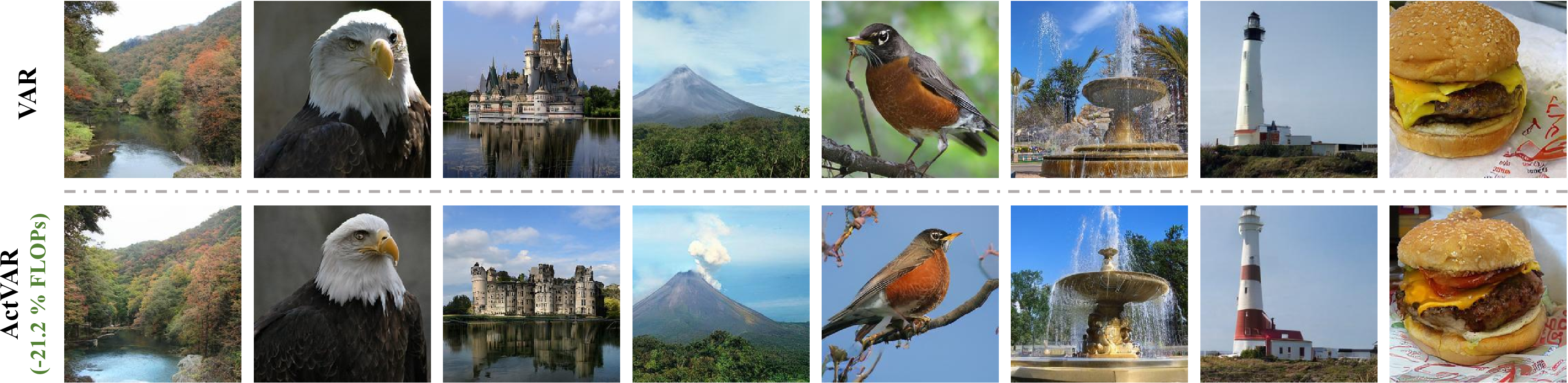} 
\caption{\textbf{Comparison between the original VAR and our proposed ActVAR on 256$\times$256 image generation.}}
\label{fig:vis}
\end{figure*}

\subsection{Layer-Adaptive Token Activation}
In addition to improving efficiency at the weight level, adaptively selecting tokens can further reduce the computational burden of VAR models, which would otherwise need to process long token sequences. As illustrated in Figure~\ref{fig:token_active}, unlike static token pruning methods, our layer-adaptive token activation retains all tokens throughout the model. At each transformer block, we dynamically select a subset of tokens with the highest optimization potential to participate in the computation, effectively reducing the overall computational cost of the model.

Unselected tokens are not discarded. Instead, they are reconsidered in subsequent layers, allowing for broader and more flexible coverage across the entire sequence. This ensures that most tokens have opportunities to be optimized at different layers. Since these unselected tokens still contain rich positional and contextual information, we reuse them to reconstruct the complete token sequence. This reconstruction preserves input-output alignment and maintains the original spatial structure of all tokens.

\textbf{Adaptive token activation.} For the input token sequence $q_{m-1} \in \mathbb{R}^L$ of the transformer block $m$, a lightweight gated selector $G(\cdot)$ is trained to predict the probabilities $p_s \in \mathbb{R}^L$ that tokens are likely to undergo significant updates in the block.
We then utilize the $\text{top-}k$ function to obtain the indicator matrix $\mathcal{I} = \{0,1\} \in \mathbb{R}^L$ representing the activations of the tokens. Depending on this indicator matrix, we extract the corresponding tokens from $q_{m-1}$ to form a compact sequence $\hat{q}_{m-1} \in \mathbb{R}^{K_t}$, which is then passed through the Attention and FFN module:
\begin{equation}
\hat{q}_m = \text{FFN}(\text{Attn}(\hat{q}_{m-1})),
\end{equation}
where $\hat{q}_m  \in \mathbb{R}^{K_t}$  denotes the updated sequence, and $K_t$ is the number of selected token.

To reconstruct the complete output sequence $q_m \in \mathbb{R}^L$, we map the updated sequence $\hat{q}_m$ into their original positions in $q_{m-1}$ based on the indicator matrix $\mathcal{I}$. The reconstruction process is defined as:
\begin{equation}
q_m^i = 
\begin{cases}
\hat{q}_m^i, & \text{if } \mathcal{I}_t^i = 1, \\
q_{m-1}^i, & \text{otherwise},
\end{cases}
\end{equation}
where $\mathcal{I}^i$ representS that the $i$-th position reuses the input token. This design enables complete knowledge reuse and preserves dimensional consistency between the input and output in VAR models.

\textbf{Optimize gated selector.} To simplify our method, when training the selector $G(\cdot)$, we adopt an optimization scheme similar to that used for training the router. We pass the entire input sequence $q_{m-1}$ through the transformer block to obtain a fully updated sequence $q_{m}^*$. We then measure the token-wise distance $\{d_i^{\text{t}}\}_{i=1}^M$ between $q_{m-1}$ and $q_m^*$ to quantify how much each token changes during the update.  Tokens exhibiting larger variations are regarded as carrying higher optimization potential. Guided by this metric, we employ the $\text{top-}k$ function to identify valuable tokens and construct a pseudo-label matrix $\mathcal{A}_t$ for gate training:
\begin{equation}
\mathcal{A}_t = \sigma(\text{top}-k(\{d_i^{\text{t}}\}_{i=1}^M, K_t)),
\end{equation}
where $M$ denotes the number of tokens in the input sequence. 

Guided by the pseudo-label matrix $\mathcal{A}_t$, the $G(\cdot)$ is trained using a distillation loss, which enforces consistency between its predicted token selections and the target patterns:
\begin{equation}
L_{dis}^t=\text{KL}(p_t \parallel \sigma(\mathcal{A}_t)).
\end{equation}

Furthermore, we employ the same load balancing loss to penalize the uneven activations among the experts. The details of our two-stage training strategy are provided in the Supplementary Materials.

\subsection{Inference and Computational Analysis}
\textbf{Inference Pipeline.} As illustrated in Figure~\ref{fig:framework}, in each transformer block $m$, the gated selector first analyzes the input token sequence $q_{m-1}$ and predicts the activation probability for each token. Based on this, it generates a binary indicator vector $\mathcal{I}$ that identifies which tokens should be updated. Using the vector $\mathcal{I}$, a compact token subset $\hat{q}_{m-1}$ is extracted from $q_{m-1}$ and passed sequentially through the Attention and the expert-based FFN. Within the FFN, the dynamic router further assigns each token to the most suitable expert network according to the routing probabilities. After processing, the updated token representations $\hat{q}_m$ are mapped back to their original positions in the sequence using $\mathcal{I}$, resulting in the reconstructed output sequence $q_m$ for block $m$. This output is then used as the input for the subsequent block in the generation process.

\textbf{FLOPs Saving Analysis.}  We consider the computational costs of dynamic routers and gated selectors in FLOPs estimation. Let $\eta$ denote the token saving rate, and $\mu$ denote the FFN weight saving rate. For a transformer block with input sequence length $L$ and hidden dimension $H$, the FLOPs can be reduced by $8 \cdot \eta \cdot L \cdot H^2 \cdot (1 + 2\mu)$. Besides, the additional computational cost introduced by the dynamic gating and routing mechanisms is $2 \cdot L \cdot H \cdot (1 + N - \eta \cdot N)$, where $N$ is the number of FFN expert networks. Therefore, the estimated FLOPs savings are computed as the difference between the reduction and the additional overhead:
\begin{equation}
\begin{split}
\sum_{i \in \mathcal{S}} \sum_{j \in \mathcal{D}} & [ 8 \cdot \eta_{i,\cdot} \cdot L_{i,j} \cdot H^2 \cdot (1 + 2\mu_{i,\cdot}) - \\
& 2 \cdot L_{i,j} \cdot H \cdot (1 + ( 1 - \eta_{i, \cdot}) \cdot N) ],
\end{split}
\end{equation}
where $\mathcal{S}$ denotes the set of steps that dynamically activate weight and token, and $\mathcal{D}$ is the set of all transformer blocks.

\begin{table*}[t]
\renewcommand{\arraystretch}{1.2}
\setlength{\tabcolsep}{4.5mm}
\footnotesize
\begin{center}
\begin{tabular}{c|c|ccc|cccc}
\hline
Type & Method  & FLOPs Saving (\%) & \#Para & \#Step & Fid ($\downarrow$) & IS ($\uparrow$) & Pre($\uparrow$) & Rec ($\uparrow$)\\
\hline
\multirow{3}*{Diff.}
& CDM        & - &   -  & 8100 & 4.88 & 158.7 & - & -\\
& DiT-L/2    & - & 458M & 250  & 5.02 & 167.2 & 0.75 & 0.57\\
& LDM-4-G    & - & 400M & 250  & 3.60 & 247.7 & - & -\\
\hline
\multirow{3}*{AR}
& VQGAN-re   & - & 1.4B & 256  & 5.20 & 280.3 & - & -\\
& ViTVQ-re   & - & 1.7B & 1024 & 3.04 & 227.4 & - & -\\
& RQTran.-re & - & 3.8B & 68   & 3.80 & 323.7 & - & -\\
\hline
\multirow{8}*{VAR}
& VAR-d16 & - & 310M & 10 & 3.30 & 274.4 & 0.84 & 0.51\\
& VAR-d20 & - & 600M & 10 & 2.57 & 302.6 & 0.83 & 0.56\\
& VAR-d24 & - & 1.0B & 10 & 2.09 & 312.9 & 0.82 & 0.59\\
& VAR-d30 & - & 2.0B & 10 & 1.92 & 323.1 & 0.82 & 0.59\\
\cline{2-9}
& ActVAR-d16 & 20.2\%  & 310M & 10 & 3.51 & 273.5 & 0.85 & 0.49\\
& ActVAR-d20 & 21.2\%  & 600M & 10 & 2.72 & 290.4 & 0.83 & 0.55\\
& ActVAR-d24 & 21.8\%  & 1.0B & 10 & 2.20 & 300.7 & 0.82 & 0.58\\
& ActVAR-d30 & 22.3\%  & 2.0B & 10 & 2.05 & 321.6 & 0.82 & 0.59\\
\hline
\end{tabular}
\end{center}
\vspace{-2mm}
\caption{\textbf{Evaluation on the ImageNet 256x256 benchmark.}}
\label{table:ablative_main}
\end{table*}

%% file: sec/4_experiment.tex
\section{Experiments}

\subsection{Experimental Setup}
\textbf{Models and Evaluation metrics.}
We implement ActVAR on multiple parameter-scale VAR models, including \texttt{VAR-d16}, \texttt{VAR-d20}, and \texttt{VAR-d30}, using the official pre-trained model's parameters as initialization. All experiments are conducted on the ImageNet $256\times256$ conditional generation benchmark~\cite{imagenet15russakovsky}. To ensure a fair comparison, we adopt the following evaluation metrics: Frechet Inception Distance (FID), Inception Score (IS), Precision, and Recall.

\textbf{Implementation Details.}
We train the ActVAR models on the ImageNet-1K~\cite{imagenet15russakovsky} dataset. The dynamic activation strategy for both FFN weights and tokens is applied during the generation steps of the final two scales. Specifically, the token and weight activation rates are set uniformly to $75\%$ in these steps. Training is conducted in two phases. In the first phase, we train the dynamic router and the gated selector while freezing all other model parameters. In the second phase, the router and selector are kept fixed, while the expert FFNs, attention modules, and classification head are fine-tuned. For both phases, we adopt a batch size of $512$ and a learning rate of $2e-4$. The AdamW optimizer is employ with $\beta_1 = 0.9$, $\beta_2 = 0.95$, and a weight decay of $0.05$. The training lasts for 2 epochs in the first phase and 10 epochs in the second. 

\subsection{Main Results}
We compare ActVAR with state-of-the-art models, including VAR models~~\cite{tian2024visual}, AR-based models~\cite{esser2021taming, yu2021vector, lee2022autoregressive}, and Diffusion-based models~\cite{ho2022cascaded, peebles2023scalable, rombach2022high}, on the ImageNet $256\times 256$ benchmark. The results, summarized in Table~\ref{table:ablative_main}, demonstrate that ActVAR consistently achieves substantial computational savings across different model scales, with only marginal degradation in image quality. Specifically, ActVAR-d16 achieves a \textbf{20.2\%} reduction in FLOPs, with only a 0.9 decrease in IS compared to VAR-d16. ActVAR-d24 reduces FLOPs by \textbf{21.8\%}, while incurring only a 0.11 increase in FID compared to the similar-sized VAR-d24. Similarly, ActVAR-d30 maintains a $\textbf{22.3\%}$ reduction in FLOPs while preserving strong performance across all evaluation metrics. These results clearly indicate that ActVAR strikes an effective balance between efficiency and fidelity, substantially lowering computational overhead while maintaining the high generation quality characteristic of autoregressive models. Furthermore, ActVAR exhibits competitive results when compared to representative AR-based and Diffusion-based models. 

We compare the images generated by VAR-d20 and ActVAR-d20, as illustrated in Figure~\ref{fig:vis}. ActVAR-d20 delivers visual fidelity comparable to VAR-d20, showing no perceptible degradation while achieving a 21.2\% reduction in FLOPs. This indicates that the dynamic expert and token selection mechanisms in ActVAR successfully allocate computation where it is needed, minimizing redundancy without sacrificing fidelity.

\begin{table}[t]
\renewcommand{\arraystretch}{1.2}
\setlength{\tabcolsep}{8pt}
\footnotesize
\begin{center}
\begin{tabular}{c|ccc}
\hline
Activation rates & FLOPs Saving (\%) & FID ($\downarrow$) & IS ($\uparrow$) \\
\hline
(50\%, 25\%; 50\%) & 43.2\% & 3.30 & 256.8 \\
(50\%, 75\%; 25\%) & 39.9\% & 3.21 & 265.1 \\
(50\%, 50\%; 75\%) & 32.8\% & 2.92 & 274.1 \\
(75\%, 50\%; 75\%) & 28.2\% & 2.89 & 277.1 \\
\hline
(75\%, 75\%; 75\%) & 21.2\%& 2.72 & 290.4 \\
\hline
\end{tabular}
\end{center}
\vspace{-2mm}
\caption{\textbf{FLOPs saving and performance with varying activation ratios ($\alpha$, $\beta$; $\gamma$)}. Here, $\alpha$ and $\beta$ denote the token activation ratios, and $\gamma$ is the activated FFN weight ratio.}
\vspace{-2mm}
\label{table:ablative_ratio}
\end{table}

\subsection{Ablation Studies}

\subsubsection{Activation Rates.}
To validate the trade-offs between computational efficiency and model performance under varying activation rates, we compare different configurations of token and FFN activation ratios, as shown in Table \ref{table:ablative_ratio}. Lower activation rates (\emph{e.g.,} $50\%$, $25\%$; $50\%$) lead to greater FLOPs savings (up to $43.2\%$) but result in degraded generation quality, as reflected by higher FID and lower IS. In contrast, higher activation rates (\emph{e.g.,} $75\%$, $75\%$; $75\%$) reduce FLOPs savings but significantly improve performance. This shows the flexibility of ActVAR in supporting a wide range of efficiency-accuracy balance via simple adjustment of activation rates.
In our study, we adopt the ($75\%$, $75\%$; $75\%$) setting, which yields the best generation quality (FID $2.72$, IS $290.4$) while still saving $21.2\%$ of FLOPs—demonstrating that high activation remains feasible with moderate efficiency gains when performance is prioritized.

\begin{table}[t]
\renewcommand{\arraystretch}{1.2}
\setlength{\tabcolsep}{10pt}
\footnotesize
\begin{center}
\begin{tabular}{c|ccc}
\hline
Module & FLOPs Saving (\%) & FID ($\downarrow$) & IS ($\uparrow$) \\
\hline
w/o DW & 14.2\% & 2.67 & 290.5 \\
w/o AT & 9.4\%  & 2.70 & 288.6 \\
\hline
w/ DW \& AT & 21.2\% & 2.72 & 290.4 \\
\hline
\end{tabular}
\end{center}
\vspace{-2mm}
\caption{\textbf{Ablation on the proposed components.} DW: Dynamic weight activation. AT: Adaptive token activation.}
\vspace{-2mm}
\label{table:components}
\end{table}

\subsubsection{The effectiveness of proposed components.}
To evaluate the contribution of each component in ActVAR, we perform ablation studies on dynamic weight activation and adaptive token activation. The experimental results are summarized in Table~\ref{table:components}. When either module is removed, the image generation quality tends to improve slightly. However, this comes at the cost of a substantial drop in FLOPs saving rates. For instance, removing dynamic weight activation increases the IS by $0.1$, but decreases the FLOPs saving rate from $21.4\%$ to $14.2\%$. Notably, excluding adaptive token activation leads to approximately a $50\%$ reduction in FLOPs savings compared to the full ActVAR model, while the FID only decreases marginally by $0.02$. These results demonstrate that both components are critical for maintaining a favorable trade-off between image generation quality and computational efficiency in ActVAR.

\begin{table}[t]
\renewcommand{\arraystretch}{1.2}
\setlength{\tabcolsep}{14pt}
\footnotesize
\begin{center}
\begin{tabular}{c|cc}
\hline
Module & FID ($\downarrow$) & IS ($\uparrow$) \\
\hline
Fixed token pruning & 3.83 & 242.4 \\
\hline
Dynamic activating token & 2.72 & 290.4\\
\hline
\end{tabular}
\end{center}
\vspace{-2mm}
\caption{\textbf{Comparison of our dynamic token activation with fixed token pruning.}}
\vspace{-2mm}
\label{table:fixed_token_pruning}
\end{table}

\subsubsection{Analysis of dynamic token activation.}
To analyze the advantages of dynamic token activation, we visualize the distribution of activated tokens across different blocks at a $16\times 16$ spatial scale. As shown in Figure~\ref{fig:token_active}, the model exhibits diverse activation patterns: some blocks tend to focus on tokens associated with the background, while others prioritize tokens corresponding to foreground objects. This block-specific behavior highlights the adaptiveness of our proposed strategy, which effectively accommodates varying token preferences across spatial regions.

Moreover, our dynamic token activation enables broader token coverage. Since the sets of tokens activated by different blocks are not identical, their union covers a large portion of the input tokens. This ensures that most tokens can still participate in the update process, even under sparse activation constraints.

To further validate this behavior, we compare our token activation with a fixed token pruning baseline. For a fair comparison, both methods retain the same number of tokens. The key difference lies in adaptability: in fixed token pruning, once the token sequence is selected during the first step, it remains static throughout inference. In contrast, our method selects tokens adaptively for each block. Shown in Table~\ref{table:fixed_token_pruning}, dynamic token activation achieves a lower FID and improves the IS by $48.0$ points compared to fixed pruning. The results not only demonstrate the effectiveness of our method but also improve its interpretability by aligning token selection with block-level semantic preferences.

\begin{table}[t]
\renewcommand{\arraystretch}{1.2}
\setlength{\tabcolsep}{14pt}
\footnotesize
\begin{center}
\begin{tabular}{c|cc}
\hline
Method & FID ($\downarrow$) & IS ($\uparrow$) \\
\hline
Static weight pruning & 3.04 & 274.2 \\
\hline
Dynamic weight activation & 2.72 & 290.4\\
\hline
\end{tabular}
\end{center}
\vspace{-2mm}
\caption{\textbf{Comparison of our dynamic weight activation with static weight pruning.}}
\vspace{-2mm}
\label{table:weight_pruning}
\end{table}

\subsubsection{Analysis of Dynamic weight activation.}
To verify the effectiveness of dynamic weight activation, we compare it with traditional static weight pruning. For fair comparison, both methods use the same weight activation rate and are trained under the same strategy. The results are shown in Table~\ref{table:weight_pruning}. Our dynamic method achieves better performance in both FID and IS. In particular, the IS of static pruning is 16.2 lower than our method, highlighting the advantage of dynamically adapting the computation pathway rather than permanently removing parameters. These results show that our approach can allocate weights more precisely for each token, achieving efficient and accurate computation. 

Furthermore, Figure~\ref{fig:weight_active} illustrates that tokens at different steps display pronounced variations in their preferences for activated expert networks. This dynamic behavior confirms that our proposed mechanism effectively captures and adapts to these variations, enabling flexible computation while maintaining generation fidelity.

%% file: sec/5_conclusion.tex
\section{Conclusion}

In this paper, we propose ActVAR, a dynamic and efficient framework for autoregressive image generation that addresses the computational challenges of next-scale prediction models like VAR. Unlike conventional static pruning approaches that permanently remove model components and risk degrading performance, ActVAR adopts a dynamic sparsity mechanism that selectively activates important computations based on the input content at each generation step. ActVAR consists of two core modules: a learnable expert router that dynamically allocates FFN sub-networks for each token, and a gated token selector that filters out less informative tokens to reduce attention overhead. This dual sparsity design concentrates computation on the most relevant regions, while a lightweight reconstruction module preserves global context and spatial alignment by reintegrating the skipped tokens.

Extensive experiments on the ImageNet $256\times256$ conditional generation benchmark demonstrate that ActVAR achieves up to 21.2\% FLOPs reduction while maintaining competitive image quality, outperforming conventional static pruning methods. We believe ActVAR offers a promising direction for accelerating autoregressive generation models without sacrificing fidelity.

%% file: sec/6_suppl.tex
\clearpage
\setcounter{page}{1}
\maketitlesupplementary

\section{Details of two-stage training strategy}
\subsection{Stage 1: train the router and selector}
\textbf{Dynamic router.} To train the routing module without direct supervision, we adopt a knowledge distillation approach. The original feedforward network serves as a teacher, while the expert networks act as student models. For each expert, we compute its mean squared error (MSE) with respect to the teacher’s output to obtain a distance vector $\{d_i^{w}\}_{i=1}^N$, reflecting how closely each expert approximates the teacher. We then select the top-$k$ closest experts and apply a softmax to produce a pseudo-label matrix:
\begin{equation}
\mathcal{A}_w = \sigma(\text{top}-k(\{-d_i^{w}\}_{i=1}^N, K_w)),
\end{equation}
which is used to supervise the routing probability $p_w$ via a distillation loss:
\begin{equation}
L_{dis}^w = \text{KL}(p_w \parallel \sigma(\mathcal{A}_w)).
\end{equation}

To prevent the router from overloading a few experts~\cite{fedus2022switch}, we follow~\cite{lepikhin2020gshard} and add a load balancing loss:
\begin{equation}
L_{bl}^w = \frac{K_w}{N}\sum_{i=1}^N\sum_{j=1}^{K_w} \mathbb{I}_j(x_i)R_j(x_i),
\end{equation}
where $\mathbb{I}_j(x_i)$ indicates whether the $j$-th expert is selected for token $x_i$, and $N$ is the number of expert networks.

\textbf{Gated selector.} We train the token selector using a strategy similar to the router. The full input sequence $q_{m-1}$ is first processed by the transformer to produce an updated output $q_m$. We then compute the token-wise distance $\{d_i^{t}\}_{i=1}^M$ between $q_{m-1}$ and $q_m$ to estimate each token's importance — tokens with larger changes are considered more informative. Based on these distances, we apply a top-$k$ operation followed by softmax to generate a pseudo-label matrix:
\begin{equation}
\mathcal{A}_t = \sigma(\text{top}-k(\{d_i^{t}\}_{i=1}^M, K_t)),
\end{equation}
which is used to supervise the selector via a Kullback–Leibler (KL) divergence loss:
\begin{equation}
L_{dis}^t = \text{KL}(p_t \parallel \sigma(\mathcal{A}_t)).
\end{equation}

We also apply the same load balancing loss to encourage activation of different tokens:
\begin{equation}
L_{bl}^t = \frac{K_t}{L}\sum_{i=1}^N\sum_{j=1}^{K_t} \mathbb{I}_j(x_i)R_j(x_i).
\end{equation}
where L is the length of the input sequence. Moreover, we retain the original classification loss $L_{cls}$ from VAR to regularize the router and selector, promoting consistency with the final predictions of the teacher model.

Therefore, the final loss formula is expressed as:
\begin{equation}
L_{stage1} = L_{cls} + \sum^D (\alpha \cdot (L_{dis}^w + L_{dis}^t) + \beta \cdot (L_{bl}^w + L_{bl}^t)),
\end{equation}
where $\alpha$ and $\beta$ are loss weights, which are set to 0.05 and 0.01, respectively, and $D$ denotes the number of transformer blocks.

\subsection{Stage 2: fine-tune the expert FFNs}
To fine-tune the expert network and other components of the student model, we apply both block-level and final output-level distillation schemes. This strategy enables the student model to adapt to the trained router and selector while effectively transferring knowledge from the teacher model.

For each transformer block, we utilize the MSE loss to align the hidden states of the student model with those of the teacher model:
\begin{equation}
L_{b} = \text{MSE}(q_m, \overline{q}_m),
\end{equation}
where $q_m$ and $\overline{q}_m$ denote the output sequences of the student and teacher models at the $m$-th block, respectively.

In addition, we adopt the KL loss to align the output distributions of the student and teacher models:
\begin{equation}
L_{f} = \text{KL}(p \parallel \overline{p}),
\end{equation}
where $p$ and $\overline{p}$ represent the predicted probability distributions of the student and teacher models, respectively.

The final loss for the second training stage is defined as:
\begin{equation}
L_{stage2} = L_{\text{cls}}+ L_{f} + \frac{1}{D} \sum^D L_{b},
\end{equation}
where $L_{cls}$ is the classification loss, and $D$ denotes the number of transformer blocks.

\begin{table}[t]
\renewcommand{\arraystretch}{1.2}
\setlength{\tabcolsep}{7pt}
\footnotesize
\begin{center}
\begin{tabular}{c|ccc}
\hline
Activation scales & FLOPs Saving (\%) & FID ($\downarrow$) & IS ($\uparrow$) \\
\hline
(7, 8)  & 8.1\% & 2.82 & 282.7 \\
\hline
(9, 10) & 21.2\% & 2.72 & 290.4 \\
\hline
\end{tabular}
\end{center}
\caption{\textbf{FLOPs saving and performance at scales ($\omega$, $\rho$)}. Here, $\omega$ and $\rho$ represent the scales where ActVAR is employed.}
\label{table:scale_step}
\end{table}

\begin{figure*}[t]
\centering
\includegraphics[width=1.0\textwidth]{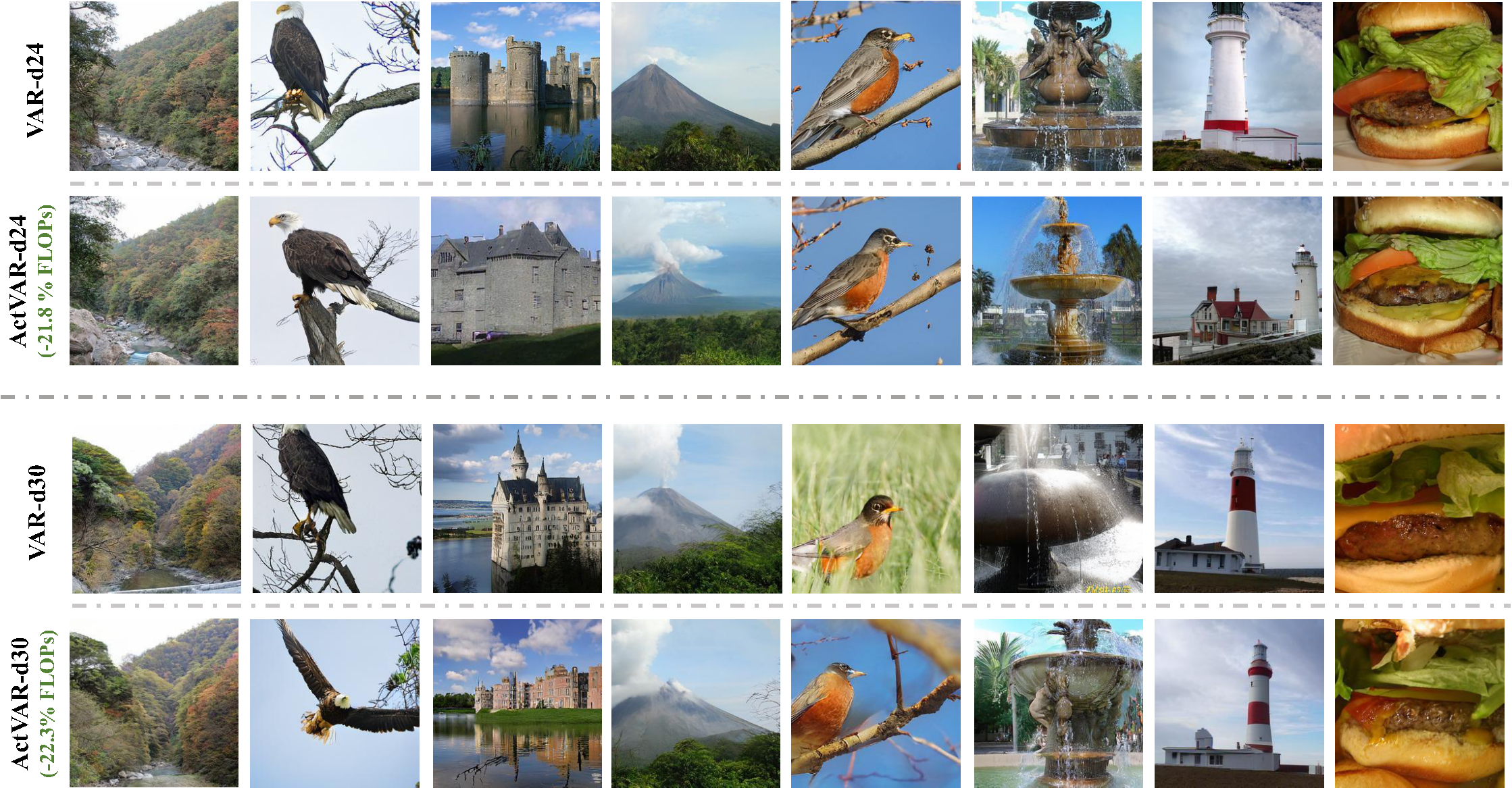} 
\vspace{-2mm}
\caption{\textbf{More visual comparisons on 256×256 image generation.}}
\label{fig:vis_more}
\end{figure*}

\section{More visual comparisons}
To verify that ActVAR maintains high generated image quality on large-scale models, we conduct the visualization on ActVAR-d24 and ActVAR-d30. As shown in the Figure~\ref{fig:vis_more}, ActVAR achieves high-quality image generation on large-scale models while reducing FLOPs by over 20\%, demonstrating its effectiveness in balancing efficiency and quality.

\section{Effects of ActVAR on different scales}
In this work, ActVAR is applied at the last two scale steps (i.e., the 9th and 10th steps). We conduct an ablation study to validate this design choice by comparing it with an alternative setting where ActVAR is applied at earlier steps (i.e., the 7th and 8th steps), while keeping the token and weight activation rates consistent. As shown in Table~\ref{table:scale_step}, applying ActVAR in earlier steps results in more noticeable performance degradation. Moreover, since the token sequences are shorter in the earlier steps, the FLOPs saving achieved under the same activation rate is also limited. Therefore, applying ActVAR in the final two steps not only achieves more effective FLOPs savings but also better preserves generation quality.

\begin{figure}[t]
\centering
\includegraphics[width=0.9\columnwidth]{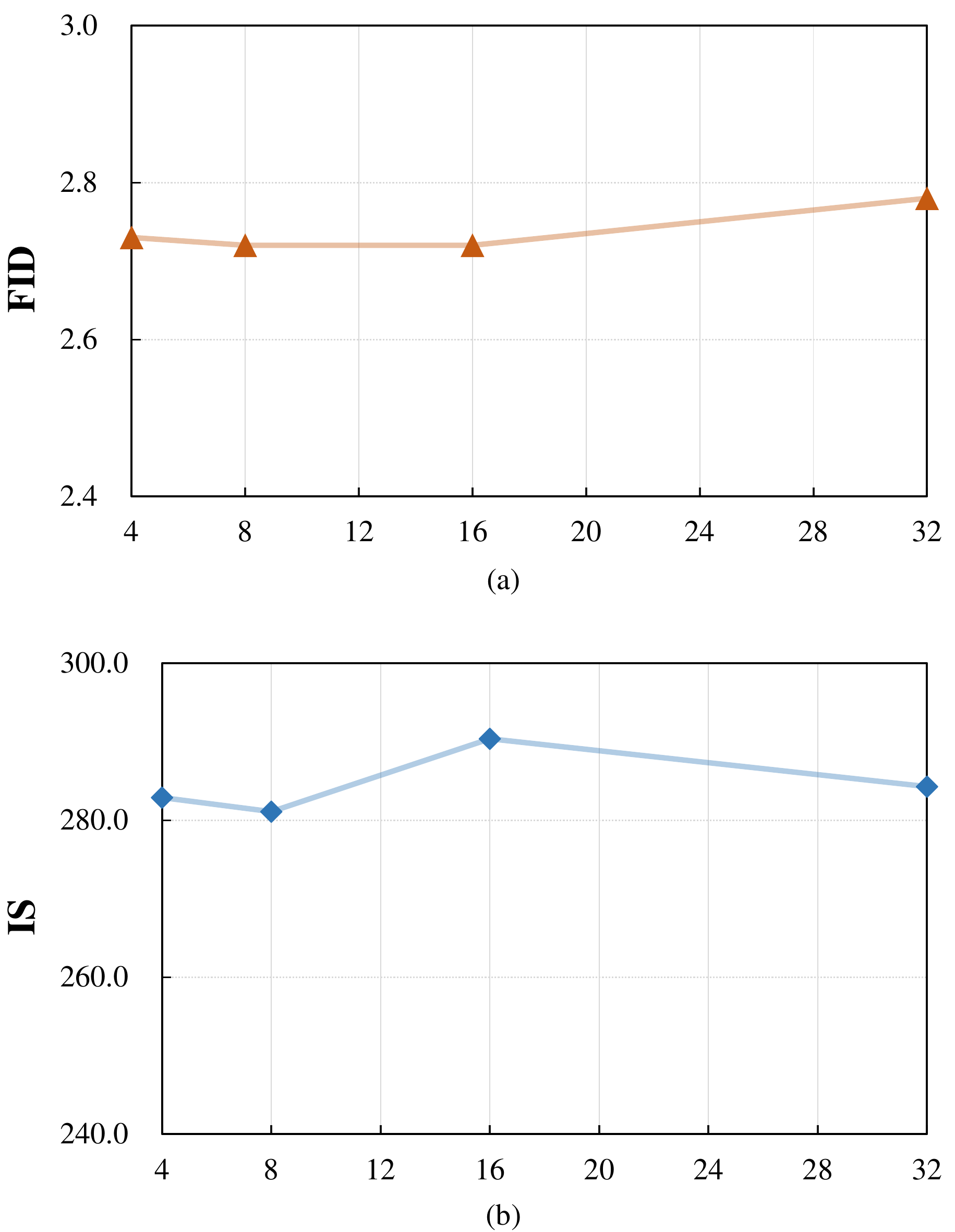}
\caption{\textbf{The impact of the number of experts on the generated image quality.}}
\label{fig:num_expert}
\end{figure}

\section{Ablation on the number of expert networks}
In this work, we decompose the original FFN into $N$ expert networks. To explore the impact of this design, we conduct ablation studies on the number of experts. As shown in the Figure~\ref{fig:num_expert}, setting a small number of experts (e.g., 4) leads to weaker performance. This is because coarse expert granularity limits the diversity of choices for input tokens. On the other hand, setting too many experts (e.g., 32) also degrades performance, as each expert has limited capacity and struggles to learn meaningful representations. Based on this trade-off, we set the number of experts to 16, which achieves an optimal balance between diversity and capacity, and yields the best performance.